\documentclass[conference]{IEEEtran}
\usepackage{cite}
\usepackage{amsmath,amssymb,amsfonts}
\usepackage{algorithmic}
\usepackage{graphicx}
\usepackage{textcomp}
\usepackage{xcolor}
\usepackage{url}
\usepackage{hyperref}
\usepackage{booktabs}   % for \toprule, \midrule, \bottomrule, \cmidrule
\usepackage{multirow}   % for \multirow
\usepackage{graphicx}

\def\BibTeX{{\rm B\kern-.05em{\sc i\kern-.025em b}\kern-.08em
    T\kern-.1667em\lower.7ex\hbox{E}\kern-.125emX}}

\begin{document}

\title{LoCC: Detection and Localization of Lip-Syncing Deepfakes via Counterfactual Frame Consistency }

% \author{Paper Id: 4088 }
\author{\IEEEauthorblockN{Soumyya Kanti Datta, Shan Jia, Siwei Lyu}
\IEEEauthorblockA{\textit{University at Buffalo, State University of New York} \\
\{soumyyak, shanjia, siweilyu\}@buffalo.edu}
% \and
% \and
% \and
% \and
% \and
}

\maketitle

\begin{abstract}
Lip-syncing deepfakes are among the most challenging forms of manipulated media because their artifacts are localized almost exclusively to the mouth region and evolve
dynamically over time. Detecting such deepfakes requires precise temporal and spatial modeling of lip motion.  In this paper, we propose LoCC, a novel detection framework that performs fine-grained detection and localization of lip-syncing deepfakes at both segment and frame levels. Unlike prior approaches that analyze videos holistically, our method evaluates whether each frame aligns with a counterfactual estimate generated from its temporal neighbors. Real videos exhibit strong and stable consistency, whereas lip-sync deepfakes introduce localized inconsistencies.  Following a teacher–student learning paradigm, our model effectively captures these frame-level discrepancies and achieves superior performance over state-of-the-art methods on multiple benchmark lip-syncing deepfake datasets, including  LAV-DF, AVDF1M, FakeAVCeleb, and KODF, and generalizes well across compression levels and datasets.  Code is available at \url{https://github.com/skrantidatta/LOCC}.

\end{abstract}


\section{Introduction}
\label{sec:intro}
The rapid advancement of generative artificial intelligence has increased the development of synthetic media, allowing the creation of highly sophisticated lip-syncing deepfakes. Lip-syncing deepfakes are digitally altered videos where a person's mouth movement is altered to correspond with a different audio track, thereby portraying the individual as speaking words they never actually spoke. When used in controlled and ethical settings, lip-syncing techniques can be useful. In the entertainment industry, they can improve audio–visual synchronization in movie dubbing and can make AI agents and avatars look more natural. Despite their benefits, lip-syncing deepfakes pose serious risks. Accessible open-source tools allow realistic deepfakes of public figures to be created with minimal expertise. Recent political misinformation and financial scam incidents demonstrate how such videos can mislead the public \cite{france24_2025}, enable fraud  \cite{ktla_2025}, damage a person’s reputation and cause lasting financial harm\cite{cnn_2024}. These highlight the high deceptive power of modern deepfakes.

\begin{figure}[t]
\centering
\includegraphics[width=\columnwidth]{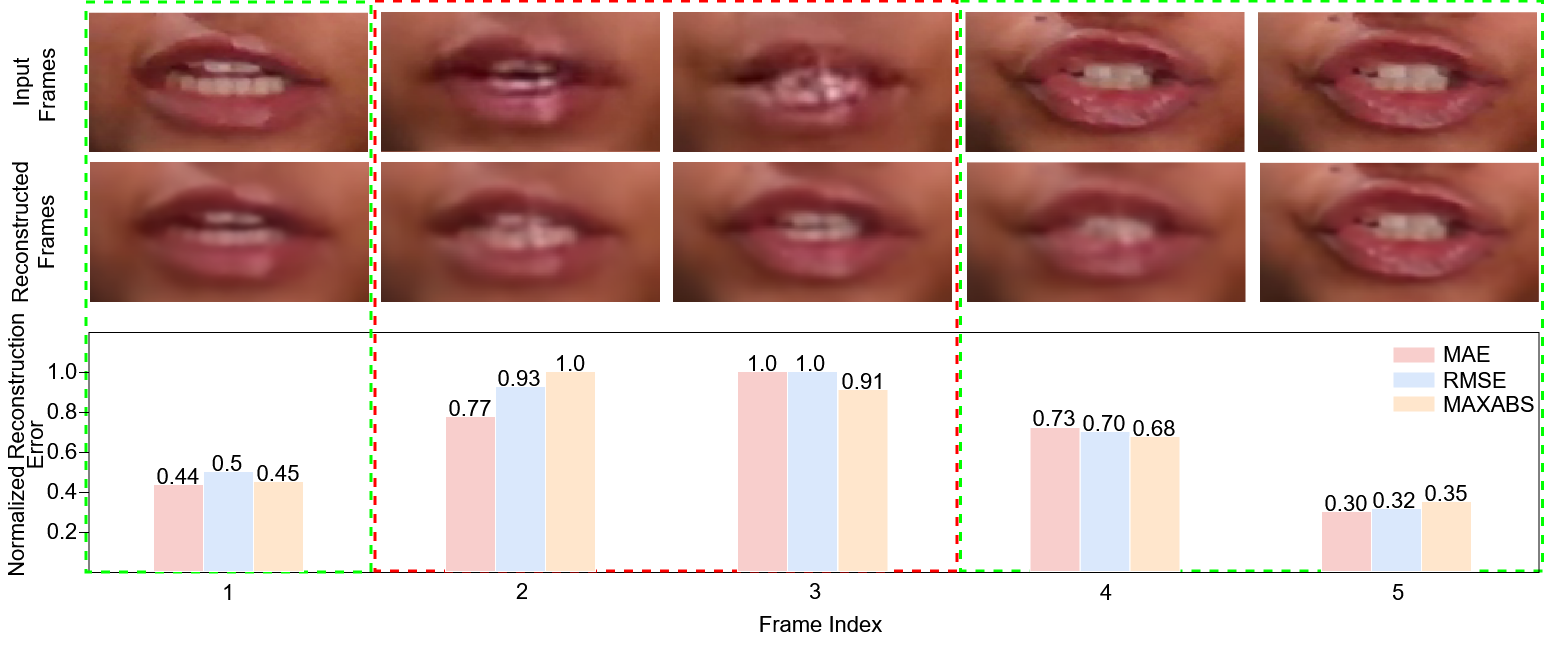}
\caption{Visualization of reconstruction-based temporal inconsistency in a partially lip-syncing deepfake video.
The top row shows consecutive input mouth frames extracted from a video, with frames highlighted in red indicating lip-syncing deepfake regions and frames in green indicating real regions. The middle row presents the reconstructed mouth frames obtained via counterfactual frame prediction using adjacent frames.
The bottom plot reports the normalized reconstruction errors for each frame, computed using MAE, RMSE, and MAXABS. 
The lip-syncing deepfake frames exhibit stronger temporal inconsistencies in the mouth region, resulting in higher reconstruction errors compared to real frames, enabling localization of deepfake segments within a video.}
\label{fig:introdiagram}
\end{figure}

Lip-sync deepfakes are particularly challenging to detect because the manipulation is typically confined to the mouth region, while the rest of the face remains visually consistent and realistic. Early lip-sync deepfake detection methods focused on spatiotemporal inconsistencies between adjacent mouth frames \cite{Lipsdontlie,Faceforensics++}, such as identifying semantic anomalies in mouth movements or spatial artifacts across facial regions. More recent approaches emphasize audio–visual coherence \cite{ics-av-deepfake,oorloff2024avff,datta2025pia}, including the analysis of temporal correlations between lip movements and audio signals or phoneme–viseme alignment. However, modern deepfakes often manipulate only short segments of a video rather than the entire sequence. As a result, methods that rely solely on temporal information struggle to detect these localized manipulations. For example, approaches that depend on audio–visual synchronization become unreliable when manipulated segments are too brief to expose cross-modal inconsistencies or when the audio is  well-aligned or noisy. Thus, detection performance decreases, resulting in a higher false positive rate.

In this work, we introduce LoCC, a three-stage framework for detecting localized lip-syncing deepfakes by analyzing counterfactual frame consistency in the mouth region. We hypothesize that, given a diffusion model trained on real mouth frames, the model can reconstruct a middle frame from its previous and next mouth frames. For real videos, the reconstructed frame closely matches the ground truth, whereas for fake videos, the reconstruction exhibits subtle inconsistencies in the mouth region relative to the ground truth. Figure~\ref{fig:introdiagram} illustrates this intuition by showing input mouth frames, their diffusion-based reconstructions from adjacent frames, and the resulting normalized reconstruction errors, where real frames exhibit low error while lip-syncing deepfakes show clear temporal inconsistencies. Here, MAE, RMSE, and MAXABS denote the mean absolute error, root mean squared error, and maximum absolute reconstruction error, respectively.
Our method consists of three components. (i) A diffusion-based reconstruction model trained on real mouth frames that learns to reconstruct the middle frame given the previous and next mouth frames. (ii) A teacher network that processes non-overlapping temporal segments and jointly models the differences between the ground-truth and reconstructed mouth frames together with the temporal relationships within each segment. This network explicitly integrates the reconstruction error signal and is trained using a novel diffusion inconsistency loss that penalizes reconstruction errors while capturing temporal structure and motion changes across frames. (iii) A student network distills the teacher’s segment-level knowledge into a frame-wise detector by assigning the teacher’s predictions to all frames within each segment. This enables efficient per-frame predictions while preserving temporal reasoning. The resulting frame-level embeddings are further aggregated using a transformer to produce video-level predictions, allowing the model to capture longer-range temporal dependencies. In summary, our work presents the following primary contributions:
\begin{itemize}
    \item We highlight that a diffusion-based reconstruction model trained on real mouth frames produces reconstructed frames that closely match the ground truth in real videos, while in lip-syncing deepfakes the reconstructed and original mouth frames exhibit subtle temporal inconsistencies that reveal manipulation artifacts.
    \item We present a three-stage detection framework that combines a diffusion-based reconstruction model, a segment-level teacher network, and a frame-wise student to capture reconstruction differences and temporal inconsistencies in the mouth region across both short and long time intervals.
    \item We propose a novel diffusion inconsistency loss that leverages reconstruction error signals to penalize reconstruction mismatches while capturing temporal structure and motion changes across frames.
    \item Our model effectively captures frame-level discrepancies and achieves strong performance on multiple benchmark lip-syncing deepfake datasets, including  FakeAVCeleb, KODF, LAV-DF, and AVDF1M, while generalizing well across different compression levels, datasets, and both localized and full-video lip-syncing deepfakes.

\end{itemize}

\begin{figure*}[ht]
    \centering
    \includegraphics[width=\textwidth]{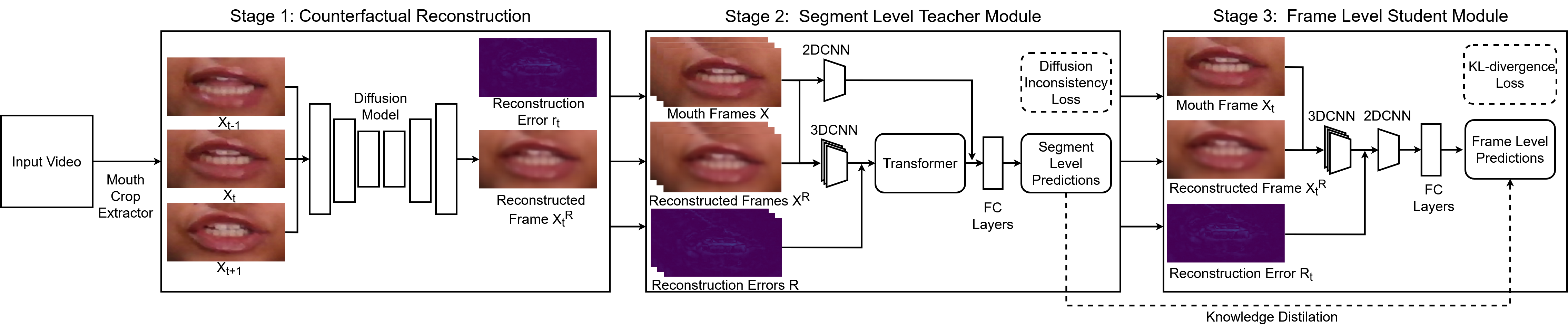}
    \caption{
    Overview of the proposed LoCC framework.
In Stage 1, given an input video, we first extract mouth-centric frames and apply a counterfactual diffusion-based reconstruction to predict the middle frame from its temporal neighbors. The reconstructed frame and its reconstruction error encode temporal consistency cues and are shared across subsequent stages. In Stage 2, short temporal segments composed of original mouth frames, reconstructed frames, and reconstruction errors are processed by a segment-level teacher to model temporal consistency and produce segment-level predictions, supervised by a diffusion inconsistency loss. In Stage 3, a lightweight frame-level student reuses the same representations to generate frame-level predictions, guided by knowledge distillation from the segment-level teacher. This design enables accurate detection of lip-syncing deepfakes.}
    \label{fig:main_model}
\end{figure*}

\vspace{-1mm}

\section{Related Work}
\noindent\textbf{Lip-syncing Generation.} As generative AI continues to advance, lip-syncing  deepfake generation methods have rapidly evolved. 
With the availability of open-source codes~\cite{wav2lip,cheng2022videoretalking} 
and applications~\cite{runwayml,sora2_openai_2025}, generating lip-syncing videos has become increasingly accessible. 
These deepfakes are difficult to detect because they affect only the mouth region.
Early lip-syncing approaches primarily relied on GAN~\cite{GAN} based architectures, 
such as Wav2Lip and VideoReTalking~\cite{wav2lip,cheng2022videoretalking}. 
More recent methods have shifted toward diffusion~\cite{dhariwal2021diffusion} based architectures, 
including Diff2Lip~\cite{mukhopadhyay2024diff2lip} and LatentSync~\cite{li2024latentsync}. 
Text-to-video generation models~\cite{runwayml,sora2_openai_2025} are also used to generate lip-syncing deepfakes and are primarily diffusion-based. These deepfakes are therefore harder to detect.

\noindent\textbf{Lip-syncing Deepfake Detection.}  As the lip-syncing deepfake generation methods are becoming more advanced, the detection techniques are improving as well. Early lip-syncing deepfake detection methods mainly focused on visual artifacts and irregular frame patterns \cite{Lipsdontlie,FTCN}. Recent methods analyze audio-visual inconsistencies \cite{AVAD,datta2025pia,oorloff2024avff} such as phoneme-viseme misalignment. In temporal deepfake localization, the goal is to identify fake segments in a video. Several methods predict the start and end boundaries of fake regions using confidence scores. Earlier temporal localization methods, such as \cite{UMMAFormer}, localize deepfake segments over time but rely on full-frame features and target general deepfakes rather than lip-syncing deepfakes. Recent methods like \cite{smeu2025circumventing} identified temporal silence bias in datasets and proposed a method which learns frame-level audio–visual consistency. \cite{ics-av-deepfake} proposed a two-stage synchronization-based audio–visual framework that learns intra and cross modal temporal consistency on real videos, to detect localized partially manipulated deepfake segments.

\vspace{-3mm}

\section{Method}
The emergence of lip-syncing deepfakes that manipulate only certain parts of a video rather than the entire video has significantly increased the difficulty of deepfake detection, as lip-syncing already alters a very small facial area, and additional variations in video resolution and lighting conditions further exacerbate this challenge. 
To address these challenges, we propose LoCC, a three-stage framework for localized lip-syncing deepfake detection using localized counterfactual frame consistency in the mouth region.
Our method is motivated by the observation that reconstructing a middle mouth frame from its neighboring frames using a diffusion model trained only on real data yields temporally coherent mouth motion and accurate reconstructions for real videos, while lip-syncing deepfakes introduce subtle reconstruction inconsistencies localized to the mouth region as shown in Fig \ref{fig:introdiagram}. Unlike prior methods such as \cite{ics-av-deepfake} and \cite{wang2023altfreezing}, which require at least 15 and 32 adjacent frames respectively to model temporal consistency, our frame wise approach requires only 1 frame combined with its counterfactual counterpart, thus significantly reducing dependence on long temporal windows. To further strengthen our frame-wise detection model, we incorporate a teacher model that processes 5 successive mouth frames, providing richer temporal context during training.  This design makes the model less sensitive to temporal shifts and reduces false positives arising from the mixed presence of real and fake segments within a video.  As a result, the model can evaluate each video frame independently to determine its authenticity.

The overall pipeline of the proposed LoCC model is depicted in Fig \ref{fig:main_model}. It comprised of 3 primary stages: (1)  Counterfactual Reconstruction module, (2) Segment-Level teacher module and (3) Frame-Level student module. In the first stage, counterfactual mouth frames are generated using a diffusion model trained exclusively on real frames. The original frames and their reconstructed counterparts are then passed to the second stage, where a teacher network operates on segments of five successive mouth frames to capture temporal context and learn reconstruction-based inconsistencies. The trained teacher subsequently guides the frame-wise student module through knowledge distillation. During inference, the student processes individual mouth frames together with their reconstructed counterparts, producing frame-level fake probability scores. In addition, the student module further incorporates a transformer-based aggregator to combine frame-level predictions into a video-level score.

\subsection{Counterfactual Reconstruction module}
In this stage, we create a diffusion-based mouth frame reconstruction model trained exclusively on real data to generate counterfactual middle frames from neighboring context frames. The model achieves 29.7 dB PSNR, 0.95 SSIM, and 0.018 MAE scores on our test dataset, indicating accurate and temporally coherent mouth reconstruction.
For each input video, we perform a structured preprocessing  by detecting the face using dlib\cite{Dlib} and extracting the corresponding mouth frames. For each frame $x_t$, we use its neighboring frames to reconstruct the middle frame $\hat{x}_t$ using our trained reconstruction model:
$
\hat{x}_t = D_r(x_{t-1}, x_{t+1}),
$
where $D_r$ denotes our diffusion-based reconstruction model. We then compute the reconstruction error $r_t = x_t - \hat{x}_t$ between the original frame and its reconstruction. Real videos exhibit low reconstruction error, whereas lip-syncing deepfakes introduce identity and temporal  inconsistencies  in the mouth region.

\subsection{Segment-Level teacher module}
In this stage, we segment the input video into segments of five successive frames, where at each time step $t$ we have the original mouth frame $x_t$, its counterfactual reconstructed frame $\hat{x}_t$, and their reconstruction error $r_t$. This module consists of two branches.
In the first branch, the model employs a 3D convolutional network~\cite{tran2015learning} to compare the original frame $x_t$ and its counterfactual counterpart $\hat{x}_t$. The resulting features are combined with the reconstruction error $r_t$ and passed through a convolutional network to produce a feature vector $\mathbf{s}_t \in \mathbb{R}^d$ for each time step. The sequence of feature vectors $\{\mathbf{s}_t\}_{t=1}^{5}$ is then fed into a transformer encoder~\cite{vaswani2017attention} that employs multi-head self-attention to learn temporal inconsistency patterns across the five successive frames.
In the second branch, we concatenate the five original frames $\{x_t\}_{t=1}^{5}$ and feed them into a 2D convolutional neural network to capture the spatial consistency features across the frames. Finally, the outputs of the two branches are fused to form a representation $\mathbf{f}$, which is then fed into a binary classifier to produce the final segment-level prediction.

\noindent\textbf{Loss Function} To supervise the teacher network, we introduce a Diffusion Inconsistency Loss (DIL) that measures inconsistency between original mouth frames and their counterfactual reconstructions over time. Given a segment of $T{=}5$ frames, the original and reconstructed sequences are $\{x_t\}_{t=1}^{T}$ and $\{\hat{x}_t\}_{t=1}^{T}$, with reconstruction error $r_t = x_t - \hat{x}_t$. Temporal differences are defined as $\Delta x_t = x_t - x_{t-1}$ and $\Delta \hat{x}_t = \hat{x}_t - \hat{x}_{t-1}$.  The DIL combines motion and reconstruction cues. We use cosine similarity between $\Delta x_t$ and $\Delta \hat{x}_t$ ($s_{\mathrm{vel}}$), temporal consistency of residual motion ($s_{\mathrm{rv}}$), reconstruction error magnitude ($s_r$), and motion mismatch magnitude ($s_{\Delta}$). These are aggregated into a consistency score $s \in (0,1)$:
\begin{equation}
s = 
\frac{w_{\mathrm{vel}} s_{\mathrm{vel}} + w_{\mathrm{rv}} s_{\mathrm{rv}} + w_r s_r + w_{\Delta} s_{\Delta}}
{w_{\mathrm{vel}} + w_{\mathrm{rv}} + w_r + w_{\Delta}}
\label{eq:dil}
\end{equation}
where $s_{\mathrm{vel}}$ denotes cosine similarity between $\Delta x_t$ and $\Delta \hat{x}_t$, $s_{\mathrm{rv}}$ denotes cosine similarity between successive residual-motion differences, $s_r$ is derived from $\lVert r_t \rVert$, and $s_{\Delta}$ is derived from $\lVert \Delta x_t - \Delta \hat{x}_t \rVert$. Magnitude-based terms are converted into similarity scores via exponential normalization. Here weights $w_{\mathrm{vel}}{=}1.0$, $w_{\mathrm{rv}}{=}0.5$, $w_r{=}0.5$, and $w_{\Delta}{=}0.5$. 
% (real $\rightarrow 1$, fake $\rightarrow 0$). 
For a real segment, the consistency score $s$ is expected to be close to $1$, while for a fake segment it should be close to $0$.
We optimize:
\begin{equation}
L_{\mathrm{DIL}} =
\frac{1}{M}\sum_{i=1}^{M}
\Big(
(1 - y_i)[-\log(s_i)] + y_i[-\log(1 - s_i)]
\Big),
\end{equation}
where $M$ is the number of segments and $y_i \in \{0,1\}$. The final teacher loss is $L_{\mathrm{teacher}} = L_{\mathrm{CE}} + \lambda L_{\mathrm{DIL}}$, where $L_{\mathrm{CE}}$ is the cross-entropy loss and $\lambda$ is the DIL weight.

\subsection{Frame-Level student module}
In this stage, we process the individual mouth frames together with their reconstructed counterparts and reconstruction error, and produce frame-level fake probability scores. The student network is trained via knowledge distillation from the segment-level teacher, enabling it to inherit temporal inconsistency cues learned from short frame sequences. For each frame, similar to our teacher module we compare it to its reconstructed counterpart using a 3D CNN to capture spatio temporal cues as well as inconsistencies. The resulting features are fused with the reconstruction error and passed through a 2D convolutional network to produce a compact frame-level embedding. A lightweight multi-layer perceptron then maps this embedding to a binary prediction for each frame.  In addition, the student incorporates a transformer-based aggregation module that combines frame-level fake probability scores into a video-level fake probability score when video-level prediction is required. 

\noindent\textbf{Loss Function:} We optimize the student module using a combination of cross-entropy loss and knowledge distillation. Given the student frame logits and teacher segment logits, we define $ \mathcal{L}_{\mathrm{student}} = (1-\alpha)\,\mathcal{L}_{\mathrm{CE}} + \alpha\,\mathcal{L}_{\mathrm{KD}},$ where $\mathcal{L}_{\mathrm{CE}}$ is the cross-entropy loss on frame labels, and $\mathcal{L}_{\mathrm{KD}}$ is the KL-divergence between the student and teacher distributions with a distillation temperature $\tau$. Here $\alpha$ is a weighting coefficient for knowledge distillation and $\tau$ is used to scale the teacher probability distribution in the KL-divergence loss to provide smoother supervision to our student module.

\begin{table}[t]
\centering
\caption{\textbf{Comparison of AP and AUC across FakeAVCeleb(in-domain) and KODF(cross-domain) Lip-syncing Deepfakes.}}
\label{tab:comparison}
\resizebox{\columnwidth}{!}{
\begin{tabular}{l|cc|cc}
\toprule
\multirow{2}{*}{Method} & \multicolumn{2}{c|}{FakeAVCeleb} & \multicolumn{2}{c}{KODF} \\
\cmidrule{2-5}
& AP & AUC & AP & AUC \\
\midrule
Intra-modal~\cite{tian2023unsupervised} & 0.94 & 0.67 & -- & -- \\
Intra-Cross-modal~\cite{tian2023unsupervised} & 0.99 & 0.96 & -- & -- \\
RealForensics~\cite{RealForensics} & 0.91 & 0.95 & 0.96 & 0.94 \\
LipForensics~\cite{Lipsdontlie} & 0.98 & 0.98 & 0.90 & 0.87 \\
Xception~\cite{Faceforensics++} & 0.90 & 0.90 & 0.77 & 0.78 \\
FTCN~\cite{FTCN} & 0.96 & 0.97 & 0.67 & 0.68 \\
AV-DFD~\cite{zhou2021joint} & 0.99 & 0.99 & 0.80 & 0.82 \\
AVAD~\cite{AVAD} & 0.94 & 0.94 & 0.88 & 0.87 \\
VQ-GAN~\cite{esser2021taming} & 0.56 & 0.53 & 0.47 & 0.46 \\
% Lipinc \cite{LIPINC} & 0.94 & 0.97 & 0.90 & 0.91 \\
AVFF~\cite{oorloff2024avff} & 0.97 & 0.99 & 0.93 & 0.96 \\
ICS-AVDF-Frozen~\cite{ics-av-deepfake} & 0.99 & 0.97 & \textbf{0.99} & \textbf{0.99}\\
\midrule
LoCC (Ours) & \textbf{0.99} & \textbf{0.99} & 0.98 & 0.98 \\
\bottomrule
\end{tabular}
}
\end{table}

\section{Experiments}
\subsection{Experimental Settings}
\noindent\textbf{Datasets:} We use real videos from FakeAVCeleb~\cite{fakeavceleb} and DeepSpeak v2.0~\cite{deepspeak} to train our counterfactual reconstruction module. We use the DeepSpeak v2.0 dataset \cite{deepspeak}, since it offers higher visual quality as reported in~\cite{datta2025pia}. Following~\cite{ics-av-deepfake}, we train and evaluate our model on the lip-syncing deepfake videos from LavDF~\cite{BA-TFD,BA-TFDplus} and AVDF1M~\cite{av1mdf} datasets to assess the localization capability of our model. To evaluate our model on fully lip-syncing deepfake videos, we train and validate it on lip-syncing videos from the FakeAVCeleb dataset, as described in~\cite{ics-av-deepfake,oorloff2024avff}. We further evaluate cross-dataset generalization on KODF~\cite{Kodf}.

\noindent\textbf{Implementation Details:} For all experiments, we resize the mouth crops to $96 \times 192$ and use the AdamW optimizer with a weight decay of $1 \times 10^{-4}$. We use a learning rate of $2 \times 10^{-4}$ for training our reconstruction module and a learning rate of $1 \times 10^{-4}$ to train both our teacher and student modules. The transformer-based video level aggregation module is trained with a learning rate of $5 \times 10^{-4}$. We set the weighting coefficient for the diffusion inconsistency loss to $\lambda = 2 \times 10^{-3}$ and  KL-divergence loss to $\alpha=0.7$. We use distillation temperature $\tau=4$. All the models are trained for 50 epochs with a batch size of 32 using PyTorch 2.9.0 with CUDA 12.6.

\noindent\textbf{Evaluation Metrics:}We evaluate our model using four widely used metrics: Intersection over Union (IoU) for temporal localization, Average Precision (AP), Area Under the Receiver Operating Characteristic Curve (AUC), and Accuracy (ACC). Percentage points are denoted as \%-pts.

\begin{table}[t]
\centering
\caption{\textbf{Temporal deepfake localization on LavDF dataset.}
Average Precision(AP\%) at different IoU thresholds. Baseline results are taken from~\cite{ics-av-deepfake}.}
\label{tab:lavdf_tfl}
\resizebox{\columnwidth}{!}{
\begin{tabular}{l|ccc}
\toprule
Method & AP@0.5 & AP@0.75 & AP@0.95 \\
\midrule
MDS~\cite{MDS} & 12.78 & 1.62 & 0.00 \\
BMN~\cite{BMN} & 24.01 & 7.61 & 0.07 \\
AVFusion~\cite{AVFusion} & 65.38 & 23.89 & 0.11 \\
BA-TFD~\cite{BA-TFD} & 76.90 & 38.50 & 0.25 \\
ActionFormer~\cite{ActionFormer} & 85.23 & 59.05 & 0.93 \\
TriDet~\cite{TriDet} & 86.33 & 70.23 & 3.05 \\
BA-TFD+~\cite{BA-TFDplus} & 96.30 & 84.96 & 4.44 \\
UMMAFormer~\cite{UMMAFormer} & \textbf{98.83} & \textbf{95.54} & \textbf{37.61} \\
ICS-AVDF-Frozen~\cite{ics-av-deepfake} & 87.40 & 66.80 & 5.72 \\
\midrule
LoCC (Ours) & 91.00 & 74.07 & 2.46 \\
\bottomrule
\end{tabular}
}
\end{table}

% ============================================================
\subsection{In-Domain and Cross-Domain Evaluation}
\label{subsec:indomain_crossdomain}

Here we train our model using the real videos and lipsyncing deepfakes from the FakeAVCeleb\cite{fakeavceleb} dataset.
We evaluate our model on the lipsyncing deepfakes from the FakeAVCeleb test split. Table~\ref{tab:comparison} reports that our model achieves 0.99 score in both AP and AUC, thus outperforming most of the baseline models and performing comparable to \cite{zhou2021joint} , \cite{oorloff2024avff} and \cite{ics-av-deepfake}. In  Table~\ref{tab:comparison}, we further evaluate cross-domain generalization on KODF~\cite{Kodf} dataset, which contains Korean-language audio-driven lip-syncing deepfakes. Using only short 5-frame segments from FakeAVCeleb and a distilled frame-wise detector, our model is able to generalize strongly and achieves 0.98 AP and 0.98 AUC score. Our model is able to outperform most of the reported baselines apart from \cite{ics-av-deepfake}, which bests our model both in terms of AP and AUC by 1\%pts.

% ============================================================
\subsection{Temporal Localization of Lip-Syncing Deepfakes}

% We first evaluate the temporal forgery localization capability of the proposed LoCC model on the LavDF dataset~\cite{BA-TFD,BA-TFDplus}. Following \cite{ics-av-deepfake}, we incorporate a boundary module \cite{BA-TFDplus} to enhance temporal localization capability of our model. We report Average Precision (AP\%) at IoU thresholds of 0.5, 0.75, and 0.95. As shown in Table~\ref{tab:lavdf_tfl}, LoCC achieves 91\%, 74.07\%, and 2.46\% at these thresholds, respectively. These results demonstrate that LoCC is competitive at moderate IoU thresholds (0.5 and 0.75) and outperforms several baseline methods, while relying only on localized mouth-region cues and a lightweight architecture of approximately 11M parameters. As noted in \cite{ics-av-deepfake}, the higher performance of some of the baseline models is likely attributable to their larger model sizes. For example, \cite{UMMAFormer} uses approximately 130M parameters, \cite{BA-TFDplus} employs around 150M parameters and \cite{ics-av-deepfake} uses about 30M parameters during pretraining and 18.8M parameters for localization.

We first evaluate the temporal forgery localization capability of the proposed LoCC model on the LavDF dataset~\cite{BA-TFD,BA-TFDplus}. Following \cite{ics-av-deepfake}, we incorporate a boundary module \cite{BA-TFDplus} to enhance temporal localization capability of our model. We report Average Precision (AP\%) at IoU thresholds of 0.5, 0.75, and 0.95. As shown in Table~\ref{tab:lavdf_tfl}, LoCC achieves 91\%, 74.07\%, and 2.46\% at these thresholds, respectively. These results demonstrate that LoCC is competitive at moderate IoU thresholds (0.5 and 0.75) and outperforms several baseline methods, while relying only on localized mouth-region cues and a lightweight architecture of approximately 11M parameters. We note that performance at stricter IoU thresholds (e.g., 0.95) is lower compared to some prior methods, likely due to their substantially larger model sizes and broader temporal modeling capacity, as also observed in \cite{ics-av-deepfake}. For example, \cite{UMMAFormer} uses approximately 130M parameters, \cite{BA-TFDplus} employs around 150M parameters and \cite{ics-av-deepfake} uses about 30M parameters during pretraining and 18.8M parameters for localization.

\begin{figure}[t]
    \centering
    \includegraphics[width=0.8\columnwidth]{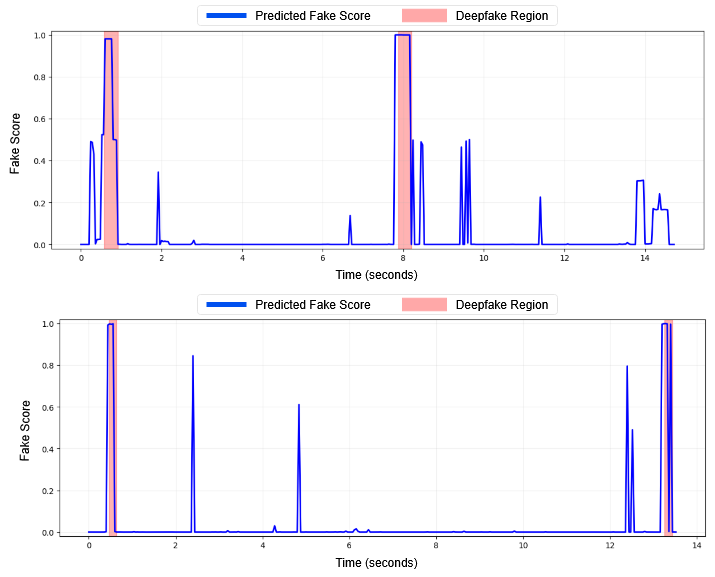}
    \caption{
    Example of Frame-wise deepfake localization on AVDF1M~\cite{av1mdf}.
    The predicted fake probability score is plotted over time.
    Real segments are shown in white and deepfake segments in red.
    }
    \label{fig:Segment}
\end{figure}

In Table~\ref{tab:avdf1m}, we further evaluate LoCC on the large-scale AV-Deepfake1M (AVDF1M) dataset~\cite{av1mdf}. Following ~\cite{ics-av-deepfake}, we report AP at IoU thresholds of 0.5, 0.75, and 0.95. For this dataset we only have 1 baseline model \cite{ics-av-deepfake} which we are able to outperform. We mainly use this dataset for ablation analysis. To further illustrate the localization capability of the proposed LoCC model on AVDF1M~\cite{av1mdf}, we visualize the per frame predictions produced by our frame wise detection model in Fig. \ref{fig:Segment}. Here the predicted fake probability score is plotted over time. Real regions are shown in white and deepfake regions in red. The results show that our model can correctly detect when a lip-syncing deepfake starts and ends, and its predictions remain stable over time instead of changing randomly. This shows the effectiveness of our model in detecting localized deepfakes.

\begin{table}[t]
\centering
\caption{\textbf{Temporal deepfake localization on AVDF1M.}
Average Precision (AP\%) at different IoU thresholds. Baseline results are taken from~\cite{ics-av-deepfake}.}
\label{tab:avdf1m}

\resizebox{\columnwidth}{!}{%
\begin{tabular}{lccc}
\toprule
Method & AP@0.5 & AP@0.75 & AP@0.95 \\
\midrule
ICS-AVDF-Frozen~\cite{ics-av-deepfake} & 23.43 & 3.48 & 0.00 \\
\midrule
\textbf{LoCC (Proposed)} & \textbf{41.22} & \textbf{18.23} & \textbf{0.17} \\
Segment-Level Teacher Only (ours)  & 30.50 & 7.91 & 0.01 \\
Frame-Wise Student Only (ours) & 33.65 & 14.55 & 0.00 \\
Segment-Level Teacher w/o DIL (ours)  & 39.41 & 15.95 & 0.06 \\
\bottomrule
\end{tabular}
}
\end{table}

\subsection{Ablation Analysis}
In Table~\ref{tab:avdf1m}, we analyze the contribution of the major components of our proposed architecture, namely the Segment-Level teacher module, Frame-Level student module, and the Diffusion Inconsistency Loss (DIL), using the AVDF1M~\cite{av1mdf} dataset. It can be observed that removal of any component of the proposed LoCC model reduces its localization capability.
For the Segment-Level Teacher Only model, we remove the Frame-Wise student module and predict probabilities for each non-overlapping segment, which are then assigned to all frames within that segment. This works well when segments contain only real or only fake frames, however when mixed segments consisting of both real and fake frames are present, localization performance degrades. As a result, performance drops by 10.72, 10.32, and 0.16~\%pts at AP@0.5, AP@0.75, and AP@0.95 respectively as compared to our proposed LoCC model. In the Frame-Wise Student Only setting, the model loses the temporal supervision provided by the teacher module, leading to performance drops of 7.57, 3.68, and 0.17~\%pts at AP@0.5, AP@0.75, and AP@0.95 respectively. In the Segment-Level Teacher w/o DIL setting, we remove the Diffusion Inconsistency Loss, thereby losing its regularizing effect. This results in performance drops of 1.81, 2.28, and 0.11~\%pts at AP@0.5, AP@0.75, and AP@0.95 respectively. From this analysis, we observe that integrating the Segment-Level teacher module with the DIL and the Frame-Wise student module provides the most robust model for detecting localized lip-syncing deepfakes.

\subsection{Conclusion}

In this paper, we proposed LoCC, a novel framework for detecting and localizing lip-syncing deepfakes by analyzing counterfactual frame consistency in mouth motion. We demonstrate that real videos exhibit strong temporal consistency between original and reconstructed mouth frames, while lip-syncing deepfakes introduce subtle inconsistencies that can be effectively exploited for detection. By combining a diffusion-based reconstruction model with a segment-level teacher guided by a diffusion inconsistency loss and a frame-wise student under a teacher–student learning paradigm, our method captures fine-grained temporal discrepancies at both segment and frame levels. Extensive experiments on multiple benchmark datasets demonstrate the effectiveness of LoCC in detecting both fully manipulated and localized lip-syncing deepfakes, achieving strong in-domain and cross-dataset performance. In future work, we aim to incorporate audio-based cues to enable detection of localized audio-only deepfakes, where visual content remains unaltered.

\bibliographystyle{IEEEbib}
\bibliography{icme2026references}

\end{document}